\documentclass{llncs}
\usepackage[russian, english]{babel}
\usepackage[utf8x]{inputenc}
\usepackage[T1]{fontenc}
\usepackage{authblk}
\usepackage{caption}
\usepackage{graphicx}
\usepackage{float}
\usepackage{makecell}
\usepackage{tabularx}
\usepackage{url}
\usepackage{booktabs}
\begin{document}
\title{Automated Detection of Non-Relevant Posts on the Russian Imageboard ``2ch'': Importance of the Choice of Word Representations}
\titlerunning{Word Embedding Models} 
%
\author{Amir Bakarov\inst{1}\inst{2} \and Olga Gureenkova\inst{1}\inst{2}}
%
\authorrunning{Bakarov and Gureenkova} 
%
\tocauthor{Amir Bakarov, Olga Gureenkova}
\institute{
Chatme AI LLC, ul. Nikolaeva 11, of. 707, Novosibirsk, Russia, \\
\and
Novosibirsk State University, Novosibirsk, Russia,\\
\email{\{a.bakarov, o.gureenkova\}@expasoft.ru}}
\maketitle              
\begin{abstract}
This study considers the problem of automated detection of non-relevant posts on Web forums and discusses the approach of resolving this problem by approximation it with the task of detection of semantic relatedness between the given post and the opening post of the forum discussion thread. The approximated task could be resolved through learning the supervised classifier with a composed word embeddings of two posts. Considering that the success in this task could be quite sensitive to the choice of word representations, we propose a comparison of the performance of different word embedding models. We train 7 models (Word2Vec, Glove, Word2Vec-f, Wang2Vec, AdaGram, FastText, Swivel), evaluate embeddings produced by them on dataset of human judgements and compare their performance on the task of non-relevant posts detection. To make the comparison, we propose a dataset of semantic relatedness with posts from one of the most popular Russian Web forums, imageboard ``2ch'', which has challenging lexical and grammatical features. 
\end{abstract}
\section{Introduction}
Currently many of the Web forums work not only as platforms for conversational entertainment but also as free sources of information in different domains of human knowledge. However, these sources are becoming significantly noised with large amounts of non-relevant posts like flame, cyber-bullying, political provocations or any other types of posts that obstruct productive discussions and interrupt convenient reading of forum thread; so, \textit{non-relevant posts could be considered as not related to the topic of the opening post of the Web forum discussion thread}. Therefore, the task of automated detection of non-relevant posts, the solution of which will allow simplifying the process of their deletion, could be approximated with the task of automated detection of \textit{semantic relatedness} (which, in this study, we consider as an existence of a common concept or a field between two linguistic units) between the given post and the opening post of the forum discussion thread. \par
In recent research studies the task of semantic relatedness detection is usually resolved by modeling semantic meaning of the matched linguistic units. This modeling is usually performed with the help of \textit{distributional semantic models}, the approaches that can represent linguistic units through dense real-valued vectors, and if the units are words, the vectors will be called \textit{word embeddings (WE)}. However, we believe that the success of such modeling is quite sensitive to the choice of word representations which vary with different word embedding models (WEM). To this end, we propose a comparison of the models in the task of semantic relatedness detection based on the approach when the vector of a posts pair could be obtained as an arithmetic mean of their non-ordered WE. So, we propose a dataset for semantic relatedness with posts from one of the most popular Russian Web forums, \textit{imageboard ``2ch''} (\selectlanguage{russian}``Двач''\selectlanguage{english}; \path{https://2ch.hk/}), containing a lot of Web slang vocabulary, misspellings, typos and abnormal grammar. But, firstly, we evaluate different WE on the dataset of word similarities to ensure that the single word representation proposed by the compared models are really different. To summarize, our main contributions are the following:
\begin{itemize}
\item Our work is the first towards a survey of the WEM applied to the textual data of Russian language, and we suggest a model for automated detection of non-relevant Web forum posts which obtained a maximum F1-score of 0.85 on our data;
\item We provide a manually annotated Russian Web slang dataset of semantic relatedness containing 2663 post pairs. 
\end{itemize}
The paper is organized as follows. Section 2 provides a survey on the related work in the given task. In Section 3 we provide description of our dataset. In Section 4 the details of the experiments are described. Section 5 covers the results of the comparison and Section 6 concludes the work. \par
\section{Related Work} \label{sec:relatedwork}
In recent years the research interest to online social media have significantly increased, and different studies have explored the Web forums from the point of natural language processing tasks like speech acts classifying \cite{qadir2011classifying}. However, we are not aware of any research in the task of post relevance detection, especially from the perspective of semantic relatedness detection of complex linguistic units (like sentences and texts) for the Russian language. But the detection of semantic relatedness itself has a broad amount of resolutions proposed by other researchers; the extensive survey of them is presented at the official web-page of \textit{Stanford Natural Language Inference Corpus} (\path{https://nlp.stanford.edu/projects/snli/}). For English most of the research of semantic relatedness/similarity were proposed as the part of \textit{*SEM shared tasks} (for example, for the task of textual semantic similarity detection \cite{agirre2013sem}); there were also some studies in the task of word similarity for Russian language as a part of \textit{RuSSE} \cite{panchenko2016human}. In this study we will also use datasets proposed on RuSSE to evaluate the word representations obtained with different WEM.  \par
\section{Dataset for semantic relatedness} \label{sec:dataset}
To propose the comparison in the task of semantic relatedness detection we created a dataset of 2663 Russian language pairs of short (up to 216 symbols) texts based on a set of posts mined from 45 different discussion threads of ``2ch'' (\textit{2ch Semantic Relatedness Dataset, 2SR}). The dataset is presented in a form of a list of triples \path{(post, op_post, is_related)} (which stands for ``single post'', ``opening post'', ``existence of semantic relatedness'') and contains human judgements about existence of semantical relatedness between the given post and the opening post in a form of binary labels; the distribution of labels in the dataset is 48\% to 52\%. 2SR notably contains a large amount of duplicates of \path{op_post} since a single opening post is associated to a large amount of posts in the structure of the Web forums, and, due to the peculiarities of the source, it is filled with misspelled, slang and obscene vocabulary. \par 
In order to collect the human judgements, three native speaking volunteers from Novosibirsk State University were invited to participate in the experiment. Each annotator was provided with the whole dataset and asked to assess the binary label to each pair choosing from options ``relatedness exists'' and ``relatedness does not exist''. To conclude the inter-annotator agreement the final label for each pair was obtained as a label marked by most of the annotators. \par
\section{Experimental setup} \label{sec:setup}
\subsection{Explored models}
For training WEM we created a corpus of 1 906 120 posts (614 707 unique words) from ``2ch''. The downloaded posts were cleared from HTML-tags, hyperlinks and non-alphabetic symbols; we also lemmatized them with \textit{pymorphy2}. \par
We set the dimensionality of word vectors to 100 (since it showed the better performance on our data across other dimensionalities); for every model we also picked the most efficient architecture based on the evaluation on our data. As a result, the following models were compared:
\begin{itemize}
\item \textbf{Word2Vec (CBOW)} \cite{mikolov2013distributed} Computation of the prediction loss of the target words from the context words. Used \textit{gensim} implementation.
\item \textbf{GloVe\footnote{\path{https://github.com/stanfordnlp/glove}}} \cite{pennington2014glove} Dimensionality reduction on the co-occurrence matrix.
\item \textbf{Word2Vec-f (CBOW)\footnote{\path{https://bitbucket.org/yoavgo/word2vecf}}} \cite{levy2014dependency} Extension of Word2Vec with the use of arbitrary context features of dependency parsing\footnote{this model was trained on a raw corpus represented in \textit{CONLL-U} format through the parsing of \textit{SyntaxNet Parsey McParseface} trained on \textit{SynTagRus}.}.
\item \textbf{Wang2Vec (Structured Skip-N-Gram)\footnote{\path{https://github.com/wlin12/wang2vec}}} \cite{ling2015two} Extension of Word2Vec with the sensitivity to the word order.
\item \textbf{AdaGram\footnote{\path{https://github.com/lopuhin/python-adagram}}} \cite{bartunov2016breaking} Extension of Word2Vec learning multiple word representations with capturing different word meanings \footnote{since AdaGram has an opportunity to predict multiple meanings for a single word, we used the most probable predicted meaning of 2 prototypes.}.
\item \textbf{FastText (CBOW) \footnote{\path{https://github.com/facebookresearch/fastText}}} \cite{bojanowski2016enriching} Extension of Word2Vec which represents words as bags of character n-grams.
\item \textbf{Swivel} \cite{shazeer2016swivel} Capturing unobserved (word, context) pairs in sub-matrices of a co-occurrence matrix. Used \textit{Tensorflow} implementation.
\end{itemize}
\subsection{Word semantic similarity}
First of all, we considered a comparison on the task of semantic similarity on three datasets of RuSSE: \textit{HJ}, \textit{RT} (test chunk) and \textit{AE} (test chunk). For each word pair of the dataset we computed the cosine distance of the embeddings associated to them, and then calculated a Spearman's correlation $p$ and an average precision score (AP) between given cosine distances and human judgements. Entries containing at least one out-of-vocabulary (OOV) word were dropped, and amount of dropped pairs consisted 27.4\% for the first dataset, 74.0\% for the second and 38.0\% for the third for Word2Vec-f models (since it uses a different vocabulary) and 5.5\%/40.9\%/9.4\% for other models. \par
\subsection{Semantic relatedness of short texts}
Secondly, for performing a comparison on the task of semantic relatedness detection, we transformed 2SR to a vector space. To obtain a single post vector, we associated a WE to each word in a single post (OOV words were not taken into account) and calculated the arithmetic mean of the obtained unorderded vectors. Then, to obtain the ``final vector'' of the pair of posts we considered three possible ways: \par
\begin{itemize}
\item Arithmetic mean of the single posts vectors (SUM);
\item Concatenation of the single posts vectors (CON);
\item Concatenation and then reducing the dimensionality twice  (we used a method of \textit{Principal Component Analysis (PCA)}).
\end{itemize} 
Then these ``final vectors'' were used as the feature matrix for learning the classifier, and the vector of labels \path{is_related} of 2SR was used as a target vector. The matrix and the vector were used as training data for the classifier which was implemented with K-Nearest Neighbors algorithm (KNN) with 3 folds and a cosine metric with the help of \textit{scikit-learn} (we also tried to use other classification algorithms and obtained lower results). We used cross-validation on the training set by 10 folds to train and to evaluate KNN. \par
The code on Python 3.5.4, 2SR, training corpus and links to the models for reproducing the experiments are available on our GitHub: \path{https://github.com/bakarov/2ch2vec}. \par
\section{Results} \label{sec:results}
\setlength{\tabcolsep}{8pt}
\begin{center}
\captionof{table}{Performance of the vectors of the compared models across different tasks. Word similarity task reports Spearman’s $p$ and AP with human annotation; semantic relatedness task reports $F1$ on different approaches of vector composing. In all cases, larger numbers indicate better performance.}
\begin{tabular}{lcccccc}
\toprule 
    Model & \multicolumn{3}{c}{Semantic Similarity} & \multicolumn{3}{c}{Semantic Relatedness, $F1$}\\

    & HJ, $p$ & RT, $AP$ & AE, $AP$
    & SUM & CON & CON+PCA  \\
    \midrule
    Word2Vec & 0.51 & 0.72 & 0.78 & 0.836 & 0.852 & 0.831 \\
    GloVe & 0.4 & 0.74 & 0.77 & 0.834 & 0.847 & 0.831 \\
    Word2Vec-f & 0.04 & 0.73 & 0.74 & 0.782 & 0.787 & 0.809 \\
    Wang2Vec & 0.41 & 0.72 & 0.78 & \textbf{0.839} & 0.85 & 0.84 \\
    AdaGram & 0.11 & 0.57 & 0.66 & 0.8 & 0.819 & 0.79 \\
    FastText & 0.44 & \textbf{0.76} & \textbf{0.79} & 0.832 & \textbf{0.854} & 0.841 \\
    Swivel & \textbf{0.52} & 0.74 & 0.76 & \textbf{0.839} & 0.851 & \textbf{0.842} \\
    \bottomrule
\end{tabular}
\end{center}
The results of the comparison on two tasks are proposed at the Table 1, and we also created plots of the learning curves of compared models which are proposed at the Fig. 1 to illustrate the process of training. The difference in values obtained for semantic similarity demonstrates that the word representations of the compared models disagree (\selectlanguage{russian}for instance, the cosine distance between words \textit{``кошка''} (cat) and \textit{``собака''} (dog) was 0.74 for FastText model and 0.62 for GloVe model)\selectlanguage{english}, since the models use different features of textual data to create the embeddings. And the difference in the embeddings leads to different performance in the task of semantic relatedness detection: the maximum interval between the scores reaches 0.05 of F1 which we consider as significant. So, Swivel and FastText are the best models for word similarity tasks, and Wang2Vec, Swivel and FastText are the best models for the semantic relatedness task. \par
\begin{figure}
    \includegraphics[width=\textwidth]{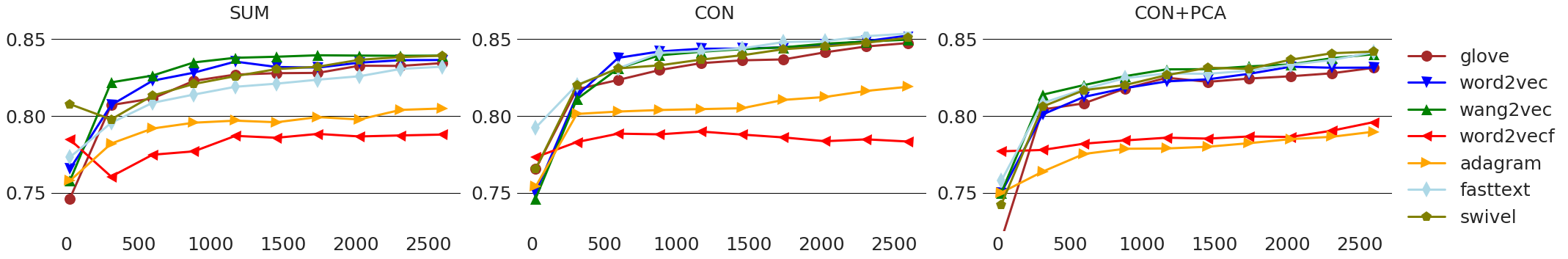}
    \centering
\captionsetup{justification=centering}
    \caption {Learning curves of KNN on cross-validation with three different options of vector composing with F1-score on Y-axis and an amount of training data on X-axis.}
\end{figure}
\section{Conclusion} \label{sec:conclusion}
The considered experiments which were illustrated with the suggested algorithm for filtering the Web forum posts confirm our hypothesis that different word representations propose different results since the nature of their embeddings varies. The best F1 on the semantic relatedness task was achieved by FastText and CON method of obtaining the pair of posts vector. However, the same model trained on 2ch corpus did not perform so good in the task of semantic similarity. It can be concluded that not only the inner algorithm of a particular WEM affects the result, but also a vocabulary of a chosen corpus. Hypothetically, the best result on a corpus should be achieved by a model which inner algorithm better reflects the human perception of semantic relatedness between different words in the particular context of this vocabulary. 
In future we plan to extend the comparison on the Web forums of other languages and propose a typological comparison of semantic drifts of the Web slang meanings in different cultures illustrating it with word representations of different languages. 
\bibliographystyle{splncs}
\bibliography{sample.bib}
\end{document}